\documentclass[letterpaper, 10 pt, conference]{ieeeconf}  % Comment this line out
\IEEEoverridecommandlockouts                              % This command is only
\usepackage[utf8]{inputenc}
\usepackage[T1]{fontenc}
\usepackage{times}
\usepackage{latexsym}
\usepackage{breqn}
\usepackage{amsmath}
\usepackage{graphicx}
\usepackage{stmaryrd} % \shortrightarrow
\usepackage{booktabs} % To thicken table lines
\usepackage{mathtools}
\usepackage{microtype}

\title{Zero-Shot Translation using Diffusion Models}

\author{Eliya Nachmani*$^{1,2}$ and Shaked Dovrat*$^{2}$% <-this % stops a space
\thanks{$^{1}$Facebook AI Research}%
\thanks{$^{2}$Tel-Aviv University}%
}

\begin{document}

\maketitle
\thispagestyle{empty}
\pagestyle{empty}

%%%%%%%%%%%%%%%%%%%%%%%%%%%%%%%%%%%%%%%%%%%%%%%%%%%%%%%%%%%%%%%%%%%%%%%%%%%%%%%%
{\let\thefootnote\relax\footnote{{*Equal contribution}}}
\begin{abstract}

In this work, we show a novel method for neural machine translation (NMT), using a denoising diffusion probabilistic model (DDPM) \cite{ho2020denoising}, adjusted for textual data, following recent advances in the field. We show that it's possible to translate sentences non-autoregressively using a diffusion model conditioned on the source sentence. We also show that our model is able to translate between pairs of languages unseen during training (zero-shot learning).

\end{abstract}

\section{Introduction}

Deep generative models can be categorized in the following categories:
(i) Flow-based models, such as Glow \cite{kingma2018glow} (ii) Autoregessive models, e.g. Transformer, for language modeling \cite{vaswani2017attention},
(iii) GAN \cite{goodfellow2014generative} based models, such as, WaveGAN \cite{donahue2018adversarial} for speech and StyleGAN \cite{karras2020analyzing} for vision application. (iv) VAE \cite{kingma2013auto} based models, e.g. VQ-VAE \cite{razavi2019generating} and NVAE \cite{vahdat2020nvae}, and (v) Diffusion Probabilistic Models \cite{sohl2015deep} such as ADM \cite{dhariwal2021diffusion}.

Diffusion probabilistic models achieve comparable and superior results to other deep generation models such as WaveGrad for speech synthesis \cite{chen2020wavegrad} and ADM for image generation \cite{dhariwal2021diffusion}.

The underline architecture of the diffusion probabilistic models is a chain of Markov latent variables. The data flows in two directions: (i) the diffusion process, and (ii) the denoising process. The denoising process is the inference process which generates the data starting from Gaussian noise. The diffusion process is the training process which learns to transform data samples into Gaussian noise.

In the seminal work of Hoogeboom et al. \cite{hoogeboom2021argmax}, a diffusion model for categorical variable was introduced. The paper shows that the original diffusion process, which is suitable for continuous data such as speech and and ordinal data such as images, can model discrete categorical data. They trained a diffusion network on the language modeling task.

In this work we propose a diffusion model for neural machine translation. Furthermore, we show that the proposed model has some capabilities of zero-shot translation. To our knowledge, we are the first to perform conditional text generation using a diffusion model.

\section{Related Work}
In \cite{sohl2015deep} Sohl-Dickstein et al. introduce the diffusion process.
The diffusion process takes the variational distribution $q(x_t|x_{t-1})$ and adds Gaussian noise at each time step where $t \in {1,...,T}$, $x_0$ is the original data point and $x_T$ is completely noise. 

In this section we will recap the multinomial diffusion process as defined by Hoogeboom et al. \cite{hoogeboom2021argmax} for categorical data. We denote $x_t$ as a 1-hot vector with $K$ categories. $x_0$ is the data point, and $q(x_t|x_{t-1})$ is the diffusion model that gradually adds a small amount of noise at each step. At $t=T$, $x_T$ is almost completely noise. The opposite direction $p(x_{t-1}|x_t)$ is a learnable distribution that denoises the data. The diffusion model is optimized with the variational bound on negative log likelihood:
\begin{multline}
    \log P(x_0) \geq E_{x_1, \ldots x_T \sim q} 
    \Big{[} \log p(x_T) \\ + \sum_{t=1}^T \log \frac{p(x_{t-1} | x_t)}{q(x_{t} | x_{t-1})} \Big{]}.
    \label{eq:diff_categorical_forward_origin}
\end{multline}
Sohl-Dickstein et al. \cite{sohl2015deep} use $x_0$ as condition and show that Eq.\ref{eq:diff_categorical_forward_origin} becomes:

\begin{multline}
    \log P(x_0) \geq  E_{q}\Big{[}\log p(x_0 | x_1)  \\ 
    - \mathrm{KL} \big{(}  q(x_T|x_0) | p(x_T) \big{)}  \\ 
    - \sum_{t=2}^T \mathrm{KL} \big{(}q(x_{t-1} | x_t, x_0) |p(x_{t-1} | x_t)\big{)} \Big{]}
\label{eq:diffusion_final_objective_kl}
\end{multline}
where $\mathrm{KL} \big{(} q(x_T | x_0) | p(x_T)\big{)} \approx 0$ if the diffusion trajectory $q$ is defined well.
The variational distribution $q(x_t|x_{t-1})$ is defined as follows:
\begin{equation}
    q(x_t|x_{t-1}) = \mathcal{C}(x_t | (1 - \beta_t) x_{t-1} + \beta_t / K )
    \label{eq:diff_categorical_forward}
\end{equation}
where $\beta_t$ is the probability to sample from the uniform distribution. Using a Markov chain property, one can get the closed form to sample $x_t$ from $x_0$:
\begin{equation}
    q(x_t | x_{0}) =  \mathcal{C}(x_t | \bar{\alpha}_t x_{0} + (1 - \bar{\alpha}_t) / K )
    \label{eq:diff_categorical_forward_x0}
\end{equation}
where $\bar{\alpha}_t$ and $\alpha_t$ are defined in the same manner as in the original DDPM \cite{ho2020denoising}, i.e. $\alpha_t = 1 - \beta_t$ and $\bar{\alpha}_t = \prod_{\tau=1}^t \alpha_\tau$. One can further relax the closed form:

\begin{equation}
    q(x_{t-1} | x_{t}, x_0) = \mathcal{C}(x_{t-1} | \tilde{\theta} / \sum_{k=1}^K \tilde{\theta}_k)
    \label{eq:q_posterior_1}
\end{equation}
where
\begin{equation}
    \tilde{\theta} = [\alpha_t x_t + (1 - \alpha_t) / K] \odot [\bar{\alpha}_{t-1} x_0 + (1 - \bar{\alpha}_{t-1}) / K ]
    \label{eq:q_posterior_2}
\end{equation}
Hoogeboom et al. \cite{hoogeboom2021argmax} predicts a probability vector for $\hat{x}_0$ from $x_t$. They parametrize $p(x_{t-1} | x_{t})$ from $q(x_{t-1} | x_t, \hat{x}_0)$, where $x_0$ is approximated with a neural network $\hat{x}_0 = \mu(x_t, t)$.
Denote 
\begin{equation}
\theta_{\mathrm{post}}(x_t, x_0) = \tilde{\theta} / \sum_{k=1}^K \tilde{\theta}_k
\end{equation}
Then, the variational lower bound Eq.\ref{eq:diffusion_final_objective_kl} becomes:
\begin{dmath}
    \log P(x_0) \geq E_{q} 
    \Big{[} \sum_k x_{0,k} \log  \hat{x}_{0,k} - \sum_{t=2}^T \mathrm{KL} \big{(} \mathcal{C}(\boldsymbol{\theta}_{\mathrm{post}}(x_t, x_0)) | \mathcal{C}(\boldsymbol{\theta}_{\mathrm{post}}(x_t, \hat{x}_0)) \Big{]}
\label{eq:loss}
\end{dmath}

It is worth to mention the work by Austin et al. \cite{austin2021structured} which improves Hoogeboom et al. \cite{hoogeboom2021argmax} by introducing corruption processes with uniform transition probabilities. They use transition matrices that mimic Gaussian kernels in continuous space and show that using different transition matrix leads to improved results in text generation.

\begin{figure}[h]
    % \centering
    \hspace{-0.9cm}
    \includegraphics[width=.55\textwidth,keepaspectratio]{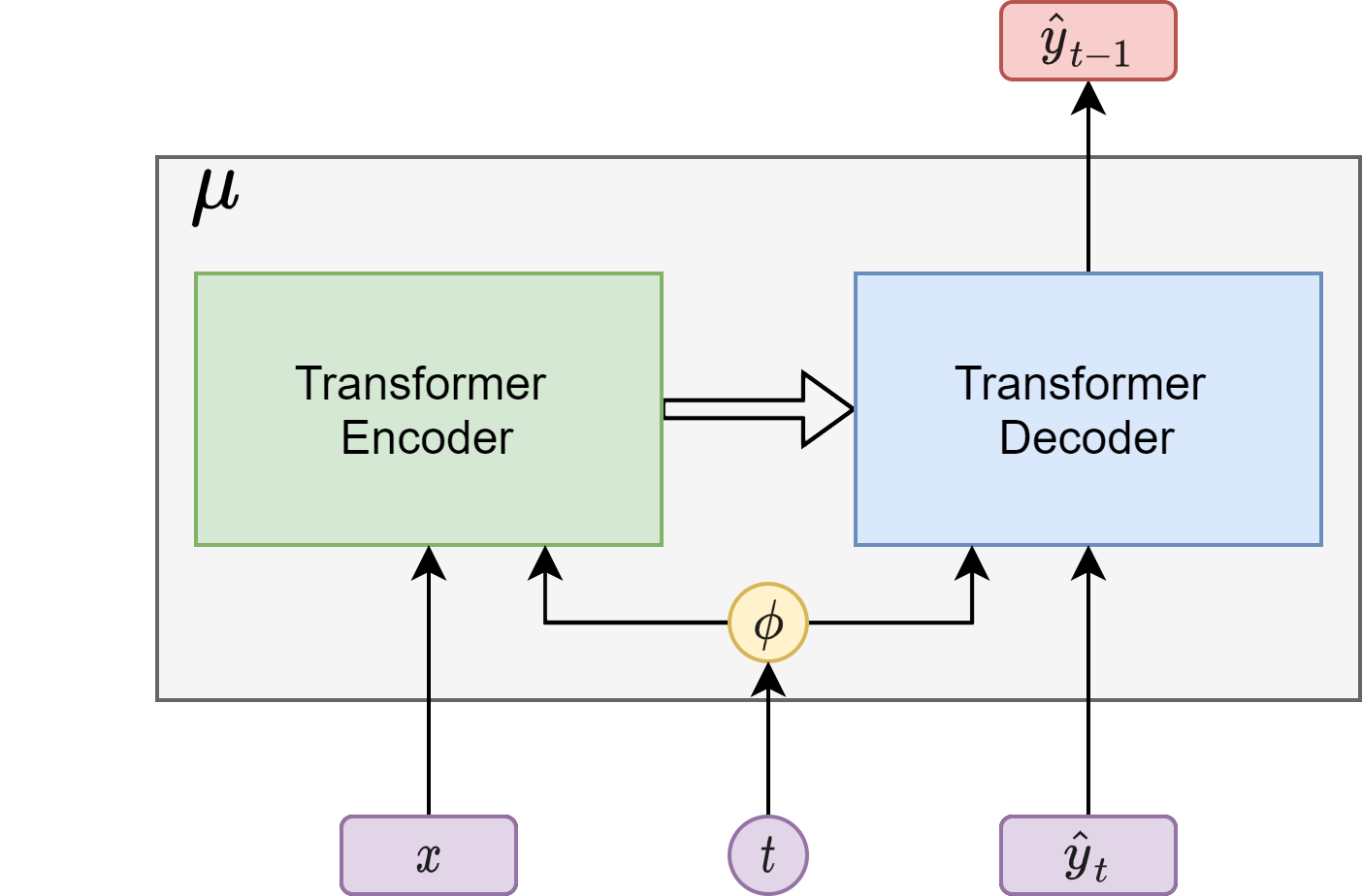}
    \caption{High level description of the proposed model. The encoder receives the source language sentence $x$, which computes outputs that are fed into the decoder using a cross attention mechanism, as customary. The decoder receives a noisy target language sentence, $\hat{y}_t$, and outputs a slightly less noisy $\hat{y}_{t-1}$. $\phi$ is the time positional encoding module, consisting of a sinusoidal positional embedding and a linear layer. Its output is added to each layer of the encoder and the decoder. This is in addition to the axial positional encoding which is built in the transformer.}
    \label{fig:arch}

\end{figure}

\section{Method}

We use a neural network $\mu$ which predicts a probability vector at each diffusion step, similar to the method used by Hoogeboom et al. \cite{hoogeboom2021argmax}. The architecture used is transformer-based with an additional time-based positional encoding. Unlike Hoogeboom et al., which performs unconditional text generation, we are interested in sentence translation, thus we add the sentence in the source language as a condition and predict probability vectors to create the sentence in the target language. Our architecture is inspired by an encoder-decoder approach, where the source-language sentence is given as input to a transformer encoder and the noisy target sentence is given as input to a transformer decoder, such that the encoder's outputs are used in a cross attention mechanism in each layer of the decoder. Unlike standard encoder-decoder systems, our method does not predict output tokens one at a time (autoregressively), but rather predicts all tokens' probabilities at each denoising step.

During training, a time step $t$ is randomly sampled, and using the noise schedule $\alpha_t, \bar{\alpha}_t$, the posteriors are calculated and a noisy target sentence $y_t$ is created, using the closed-form formula for the uniform noise \ref{eq:q_posterior_1}. Then, a forward pass is performed to predict $\hat{y}_{t-1}$ using our neural network: $\hat{y}_{t-1} = \mu(y_t, x, t)$, which in turn is used to calculate the loss function. In this notation $x$s are source sentences and $y$s are target sentences.

During inference, we start $y_T$ from random uniform noise and iteratively run $\mu$ on it $T$ times to get $\hat{y}_0$.

\noindent{\bf Data Processing\quad}
The diffusion model requires inputs to be of fixed length. Thus, we use padding and truncation of all sentences to a fixed length $L$. Sentences are padded with a special token $[PAD]$. 

Two special language tokens are added to each input sentence, the first indicates the source language and the second indicates the target language. This is used to accelerate convergence and to allow for zero-shot learning, given pairs of languages unmet during training.

\section{Experiment Setup}

\subsection{Datasets}
We used three datasets in total, all of which are from WMT. We trained our net on WMT14 \cite{bojar2014findings} DE-EN and WMT14 FR-EN jointly, and from both directions, meaning German to English, English to German, French to English and English to French. We downsampled the larger French dataset in each epoch in order to have the same number of German-English and French-English samples.

Lastly, we used the WMT19 \cite{barrault2019findings} DE-FR for evaluation only, in order to test the method's zero-shot learning performance.

\subsection{Evaluation Metrics}
We used common machine translation evaluation metrics: Corpus level BLEU, SacreBLEU, TER and chrF. We used the official SacreBLEU implementation \cite{post-2018-call} with default parameters.

\subsection{Baselines}
Current state-of-the-art results for the WMT14 translation tasks perform $\sim35$ BLEU and SacreBLEU for the German-English translation and $\sim45$ for the French-English translation. All of these methods use some type of a transformer with autoregressive decoding, use very large models, and some use extra data.

\subsection{Hyperparameter Settings}
We used the ADAM optimizer \cite{kingma2014adam}, and tuned the learning rate, batch size and gamma parameters. Eventually, we used a learning rate of $5e-4$, gamma value of $0.9$ and batch size of $512$. We also tried $1000$ versus $4000$ diffusion steps, and didn't see a major difference. We also experimented with the number of transformer layers. Our best model has $12$ transformer layers. Other parameters are $16$ attention heads and hidden dimension of $512$. 

\noindent{\bf Tokenization\quad}
A Tokenizer is trained on data from all three languages, using the WordPiece method with normalization similar to BERT \cite{devlin2018bert} (NFD Unicode, followed by Lowercase and StripAccents). Whitespace pre-tokenization is also used. The vocabulary size is a hyper parameter $K=|V|$, which we tuned.

\noindent{\bf Vocabulary Size\quad}
The vocabulary size is an important hyper parameter since it determines the dimension of the space the diffusion model needs to predict, and it changes how the noise probability is distributed (since the constant probability of change is distributed to a different number of tokens). Hoogeboom et. al. \cite{hoogeboom2021argmax} worked with 27 and 256 categories for the $text8$ and $enwik8$ datasets, respectively. The fact that no tokenization was used hinted us that perhaps large a vocabulary size doesn't work well with this method. However, austin et. al. \cite{austin2021structured} was able to train a network with 8192 categories, using a slightly different method. Furthermore, we know of the importance of tokenization in complicated tasks such as translation. Therefore, we decided to try different vocabulary sizes and see how sensitive the method is to it.

\begin{table}[]
\centering
\begin{tabular}{l|cccc}
\toprule
\textbf{Tasks}           & BLEU   & SacreBLEU  & TER   & chrF  \\
\midrule
DE$\shortrightarrow$EN   & 7.17   & 8.13       & 93.1   & 34.7   \\
EN$\shortrightarrow$DE   & 3.54   & 4.54       & 102.3  & 33.5  \\
FR$\shortrightarrow$EN   & 8.62   & 9.93       & 88.4   & 37.8   \\
EN$\shortrightarrow$FR   & 7.56   & 9.02       & 90.7   & 37.5 \\
\midrule
DE$\shortrightarrow$FR   & 4.17   & 5.06       & 94.7   & 31.4   \\
FR$\shortrightarrow$DE   & 2.96   & 4.04       & 98.1   & 31.4   \\
\bottomrule
\end{tabular}
\caption{Results for the different translation tasks. The first four rows are of supervised tasks, and the last two rows are for zero-shot tasks, i.e. pairs of languages unmet during training.}
\label{tab:results}

\end{table}

\begin{table}[]
\centering

\setlength\tabcolsep{4pt}
\begin{tabular}{l|cccc}
\toprule
\textbf{V. Size}  & DE$\shortrightarrow$EN & EN$\shortrightarrow$DE & FR$\shortrightarrow$EN &  EN$\shortrightarrow$FR  \\
\midrule
1024                                & 5.60            & 3.18         & 7.23            & 6.73               \\
2048                                & 7.92            & 4.42         & 9.83            & 8.99            \\
\textbf{4096}                       & \textbf{8.13}   & \textbf{4.54}     & \textbf{9.93}   & \textbf{9.02}              \\
8192                                & 6.76            & 4.00         & 8.89            & 7.75             \\
\bottomrule
\end{tabular}
\caption{SacreBLEU results with different vocabulary sizes. We see that $K=4096$ gives the best results, which is not on the edge of the values chosen. This indicated its a sweet spot for the vocabulary size tradeoff.}
\label{tab:vocab-size}

\end{table}

\begin{table*}[h]
\centering

\begin{tabular}{l|lcc}
\toprule
\textbf{Sample}   &  \textbf{Sentence}     &              \textbf{Lang.}    & \textbf{SacreBLEU} \\
\midrule
1st Input          & je sais qu'il voudrait une garantie de quatre ans.                                & FR   & -   \\
1st Reference      & i know he would like a four - year guarantee.                                     & EN   & -   \\
1st Prediction     & i know he need a guarantee for four years.                                        & EN & 17.47 \\
\midrule
2st Input          & \vtop{\hbox{\strut the ecb's sole mandate has always revolved around}
                           \hbox{\strut inflation, therefore mario draghi and his team have all}
                           \hbox{\strut the more reason to take action at their meeting next week.}}   & EN & -     \\
2st Reference      & \vtop{\hbox{\strut le mandat unique de la bce a toujours porte sur l'inflation,}
                           \hbox{\strut donc mario draghi et son equipe ont davantage de raisons}
                           \hbox{\strut d'agir lors de la reunion de la semaine prochaine.}}           & FR & -     \\
2st Prediction     & \vtop{\hbox{\strut lle systeme unique unique de la bce est derriere le trend}
                           \hbox{\strut en phoque, afin ou mario draghi et son mont sont plus}
                           \hbox{\strut justifies de prendre en contact a la session prochaine.}}      & FR & 17.10 \\
\midrule
3rd  Input         & \vtop{\hbox{\strut zwei kinder haben in uruguay den mord eines }
                           \hbox{\strut 11 - jahrigen eingestanden.}}                  & DE & -     \\
3rd Reference      & \vtop{\hbox{\strut two children have confessed to the murder of an }
                           \hbox{\strut 11 - year - old in uruguay.}} & EN & -   \\
3rd Prediction     & \vtop{\hbox{\strut they had spent a 118'old increased blood in uruguay }
                           \hbox{\strut about alleged murders abandone treating two children.}} & EN & 6.84 \\
\midrule
4th  Input         & town council delighted with solid budget & EN & -     \\
4th Reference      & gemeinderat freut sich uber soliden haushalt & DE & -   \\
4th Prediction     & der stadtrat erfullt einen beliebten haushalt & DE & 8.12 \\
                           
\bottomrule
\end{tabular}
\caption{Randomly selected samples from our model. 2nd sample shows a relatively good translation for a long sentence, and the 3rd sample shows a failed translation for a seemingly easier sentence.}
\label{tab:samples}
\end{table*}

\section{Results}

Results for the different translation tasks are depicted in Table \ref{tab:results}. The results are unsatisfactory, implying the method is currently not suitable for the translation task. Results for the zero-shot translation tasks (WMT19) show that some generalization to unmet pairs of languages was possible, but because the overall performance of the system is low, it is hard to estimate if the method transforms well to zero-shot learning.

Results for the vocabulary size tuning is depicted in Table \ref{tab:vocab-size}, suggesting a vocabulary of size $K=4096$ is closest to the optimal value in this case.

Qualitatively speaking, results quality vary, and overall we see an expected correlation between the difficulty of inputs and the quality of the translation. Nonetheless, some observations are hard to explain, such as relatively good translations for seemingly hard sentences and relatively bad translations for seemingly easy sentences. Table \ref{tab:samples} shows four randomly selected samples from the test set, one from each task (ordered pair of languages).

\section{Discussion}

\subsection{Learning the Transition Matrix}
One idea we had was to learn the transition matrices that determine the probabilities of noise changing one token to another. In the described "vanilla" implementation, all probabilities of change are uniform. Diffusion models for continuous or ordinal data use Gaussian noise, which gives higher probabilities to small changes, resulting in a much easier learning ground for the denoising optimization procedure. This advantage is lost when using the uniform distribution for categorical data. Austin et. al. \cite{austin2021structured} was able to improve on that by using non-uniform noise distributions.

Following this idea, we aimed to learn the noise distribution jointly with the diffusion model. Later we found out it was infeasible, since the learning procedure uses pre-computed powers of the transition matrix to enable fast learning. Specifically, for each training iteration at some $t$, this would require the computation of the $t^{th}$ power of a $K\times K$ matrix, where $K\sim 2^11$ and $t\sim 1000$. This makes the technique infeasible.

\subsection{Conclusions}
In this work, we tried to solve a thoroughly researched NLP task, MNT, using a recent and very promising method, DDPMs, for the first time (to our knowledge). This method has the potential to generate text with high performance in a non-autoregressive way.

Although DDPMs achieve state-of-the-art results in generating both continuous and ordinal data, it is yet to show competing results for categorical data such as text. We hoped to show that it can give reasonable results for non-autoregressive translation. 

% \clearpage
\bibliographystyle{IEEEtran}
\bibliography{acl2020}

\begin{thebibliography}{10}
\providecommand{\url}[1]{#1}
\csname url@samestyle\endcsname
\providecommand{\newblock}{\relax}
\providecommand{\bibinfo}[2]{#2}
\providecommand{\BIBentrySTDinterwordspacing}{\spaceskip=0pt\relax}
\providecommand{\BIBentryALTinterwordstretchfactor}{4}
\providecommand{\BIBentryALTinterwordspacing}{\spaceskip=\fontdimen2\font plus
\BIBentryALTinterwordstretchfactor\fontdimen3\font minus
  \fontdimen4\font\relax}
\providecommand{\BIBforeignlanguage}[2]{{%
\expandafter\ifx\csname l@#1\endcsname\relax
\typeout{** WARNING: IEEEtran.bst: No hyphenation pattern has been}%
\typeout{** loaded for the language `#1'. Using the pattern for}%
\typeout{** the default language instead.}%
\else
\language=\csname l@#1\endcsname
\fi
#2}}
\providecommand{\BIBdecl}{\relax}
\BIBdecl

\bibitem{ho2020denoising}
J.~Ho, A.~Jain, and P.~Abbeel, ``Denoising diffusion probabilistic models,''
  \emph{arXiv preprint arXiv:2006.11239}, 2020.

\bibitem{kingma2018glow}
D.~P. Kingma and P.~Dhariwal, ``Glow: Generative flow with invertible 1x1
  convolutions,'' \emph{arXiv preprint arXiv:1807.03039}, 2018.

\bibitem{vaswani2017attention}
A.~Vaswani, N.~Shazeer, N.~Parmar, J.~Uszkoreit, L.~Jones, A.~N. Gomez,
  {\L}.~Kaiser, and I.~Polosukhin, ``Attention is all you need,'' in
  \emph{Advances in neural information processing systems}, 2017, pp.
  5998--6008.

\bibitem{goodfellow2014generative}
I.~J. Goodfellow, J.~Pouget-Abadie, M.~Mirza, B.~Xu, D.~Warde-Farley, S.~Ozair,
  A.~Courville, and Y.~Bengio, ``Generative adversarial networks,'' \emph{arXiv
  preprint arXiv:1406.2661}, 2014.

\bibitem{donahue2018adversarial}
C.~Donahue, J.~McAuley, and M.~Puckette, ``Adversarial audio synthesis,''
  \emph{arXiv preprint arXiv:1802.04208}, 2018.

\bibitem{karras2020analyzing}
T.~Karras, S.~Laine, M.~Aittala, J.~Hellsten, J.~Lehtinen, and T.~Aila,
  ``Analyzing and improving the image quality of stylegan,'' in
  \emph{Proceedings of the IEEE/CVF Conference on Computer Vision and Pattern
  Recognition}, 2020, pp. 8110--8119.

\bibitem{kingma2013auto}
D.~P. Kingma and M.~Welling, ``Auto-encoding variational bayes,'' \emph{arXiv
  preprint arXiv:1312.6114}, 2013.

\bibitem{razavi2019generating}
A.~Razavi, A.~v.~d. Oord, and O.~Vinyals, ``Generating diverse high-fidelity
  images with vq-vae-2,'' \emph{arXiv preprint arXiv:1906.00446}, 2019.

\bibitem{vahdat2020nvae}
A.~Vahdat and J.~Kautz, ``Nvae: A deep hierarchical variational autoencoder,''
  \emph{arXiv preprint arXiv:2007.03898}, 2020.

\bibitem{sohl2015deep}
J.~Sohl-Dickstein, E.~Weiss, N.~Maheswaranathan, and S.~Ganguli, ``Deep
  unsupervised learning using nonequilibrium thermodynamics,'' in
  \emph{International Conference on Machine Learning}.\hskip 1em plus 0.5em
  minus 0.4em\relax PMLR, 2015, pp. 2256--2265.

\bibitem{dhariwal2021diffusion}
P.~Dhariwal and A.~Nichol, ``Diffusion models beat gans on image synthesis,''
  \emph{arXiv preprint arXiv:2105.05233}, 2021.

\bibitem{chen2020wavegrad}
N.~Chen, Y.~Zhang, H.~Zen, R.~J. Weiss, M.~Norouzi, and W.~Chan, ``Wavegrad:
  Estimating gradients for waveform generation,'' \emph{arXiv preprint
  arXiv:2009.00713}, 2020.

\bibitem{hoogeboom2021argmax}
E.~Hoogeboom, D.~Nielsen, P.~Jaini, P.~Forr{\'e}, and M.~Welling, ``Argmax
  flows and multinomial diffusion: Towards non-autoregressive language
  models,'' \emph{arXiv preprint arXiv:2102.05379}, 2021.

\bibitem{austin2021structured}
J.~Austin, D.~Johnson, J.~Ho, D.~Tarlow, and R.~v.~d. Berg, ``Structured
  denoising diffusion models in discrete state-spaces,'' \emph{arXiv preprint
  arXiv:2107.03006}, 2021.

\bibitem{bojar2014findings}
O.~Bojar, C.~Buck, C.~Federmann, B.~Haddow, P.~Koehn, J.~Leveling, C.~Monz,
  P.~Pecina, M.~Post, H.~Saint-Amand \emph{et~al.}, ``Findings of the 2014
  workshop on statistical machine translation,'' in \emph{Proceedings of the
  ninth workshop on statistical machine translation}, 2014, pp. 12--58.

\bibitem{barrault2019findings}
L.~Barrault, O.~Bojar, M.~R. Costa-Jussa, C.~Federmann, M.~Fishel, Y.~Graham,
  B.~Haddow, M.~Huck, P.~Koehn, S.~Malmasi \emph{et~al.}, ``Findings of the
  2019 conference on machine translation (wmt19),'' in \emph{Proceedings of the
  Fourth Conference on Machine Translation (Volume 2: Shared Task Papers, Day
  1)}, 2019, pp. 1--61.

\bibitem{post-2018-call}
\BIBentryALTinterwordspacing
M.~Post, ``A call for clarity in reporting {BLEU} scores,'' in
  \emph{Proceedings of the Third Conference on Machine Translation: Research
  Papers}.\hskip 1em plus 0.5em minus 0.4em\relax Belgium, Brussels:
  Association for Computational Linguistics, Oct. 2018, pp. 186--191. [Online].
  Available: \url{https://www.aclweb.org/anthology/W18-6319}
\BIBentrySTDinterwordspacing

\bibitem{kingma2014adam}
D.~P. Kingma and J.~Ba, ``Adam: A method for stochastic optimization,''
  \emph{arXiv preprint arXiv:1412.6980}, 2014.

\bibitem{devlin2018bert}
J.~Devlin, M.-W. Chang, K.~Lee, and K.~Toutanova, ``Bert: Pre-training of deep
  bidirectional transformers for language understanding,'' \emph{arXiv preprint
  arXiv:1810.04805}, 2018.

\end{thebibliography}

\end{document}